%% file: main.tex
\pdfoutput=1

\documentclass[11pt]{article}

\usepackage[]{acl}

\usepackage{times}
\usepackage{latexsym}

\usepackage[T1]{fontenc}

\usepackage[utf8]{inputenc}

\usepackage{microtype}

\usepackage{booktabs}
\usepackage[final]{graphicx}
\usepackage{supertabular}
\usepackage{svg}

\title{Are Fairy Tales Fair? Analyzing Gender Bias in Temporal Narrative Event Chains of Children's Fairy Tales}

\author{Paulina Toro Isaza$^1$, Guangxuan Xu$^1$\, Akintoye Oloko$^1$\\ 
\textbf{Yufang Hou$^1$, Nanyun Peng$^2$,   Dakuo Wang$^3$} \\
  $^1$ IBM Research \quad
  $^2$University of California Los Angeles 
  $^3$Northeastern University  \quad \\
  \small \tt \{ptoroisaza, gx.xu\}@ibm.com \quad
  \small \tt \href{mailto:yhou@ie.ibm.com}{yhou@ie.ibm.com} \\
  \small \tt \href{mailto:violetpeng@cs.ucla.edu}{violetpeng@cs.ucla.edu}
  \small \tt \href{mailto:d.wang@northeastern.edu}{d.wang@northeastern.edu}\\
}

\begin{document}
\maketitle
\begin{abstract}
\input{0_abstract}

\end{abstract}

\section{Introduction}
\label{sec:intro}
\input{1_introduction}

\section{Related Work}
\label{sec:related_work}
\input{2_related_work}

\section{Data Collection}
\label{sec:data}
\input{3_data}

\section{Event Type Annotation Scheme}
\label{sec:annotations}
\input{4_annotations}

\section{Analysis Methods}
\label{sec:methods}
\input{5_methods}

\section{Analysis Results}
\label{sec:discussion}
\input{6_analysis}

\section{Conclusion and Future Work}
\label{sec:conclusion}
\input{7_conclusion}

\section*{Limitations}
Our analysis is primarily limited by the accuracy of underlying NLP models used in our character event extraction pipeline. For example, BookNLP does not cluster nominal mentions of characters ("the girl") with the corresponding character names ("Cinderella"). This results in character event chains that do not account for all of the character's actual events. Using AllenNLP to extract all action verbs in a sentence as the event triggers meant that not all of our events were on the same dimension: some events were intended or thought of, while others actually happened. Additionally, narrative events that are described in ways beyond just action verbs are not extracted. (For example, the event of a kidnapping might be described as two separate actions: a character picking up another character and running away.) Our salient event identification algorithm might also filter out many events of analytic interest. 
Both characters whose gender are not specified in the story or who are gender-less are classified as ``unknown''.  There is no explicit way to extract non-binary characters as models tend to label uses of the pronoun "them" as plural. Thus, the current implementation is limited to comparisons of female and male characters which perpetuates a gender binary. 

Our use of bootstrapping to calculate confidence intervals and determine statistical significance is valid under the assumption that the original FairtytaleQA sample is representative of all fairy tales. As the sample was collected only from  popular open-source stories, this assumption may not hold. 

Lastly, bias exists beyond just gender groups and gender itself intersects with other social groups. We plan on expanding this component to include attributes such as race and ethnicity, age, and socioeconomic class. The cultural comparisons and overall analyses were too limited as the FairytaleQA dataset is very Eurocentric with most fairy-tales coming from Northern and Western Europe (Table \ref{tbl:cultures} in \ref{a:3}. Only some stories income from East Asian, Southern European, or indigenous North American cultures. Meanwhile, almost no fairy-tales are included from South America, the Middle East, Africa, South Asia, or South East Asia. Unfortunately, after considering the break down of event chains by gender and culture, the samples were too small to observe robust trends. 

\section*{Ethics Statement}

The goal of this analysis was to surface potential gender bias in story texts in new ways that were previously impossible due to the manual effort and time involved. We hope that the results will extend and deepen the analysis and discussion within the context of the rich body of work in the social sciences and humanities. We make the normative assumption that any substantial, measured numerical difference between two groups is indicative of bias within a story. We are aware that numerical measures of bias can be used to obfuscate nuance or wave away concerns of harmful representation. We do not intend for our analyses to replace qualitative analyses of stories, but rather supplement existing bias analysis frameworks, tools, and literature.

\bibliography{anthology,custom}

\appendix
\onecolumn
\section{Appendix}
\label{sec:appendix}

\input{8_appendix}

\end{document}

%% file: 0_abstract.tex
Social biases and stereotypes are embedded in our culture in part through their presence in our stories, as evidenced by the rich history of humanities and social science literature analyzing such biases in children stories. Because these analyses are often conducted manually and at a small scale, such investigations can benefit from the use of more recent natural language processing methods that examine social bias in models and data corpora. Our work joins this interdisciplinary effort and makes a unique contribution by taking into account the event narrative structures when analyzing the social bias of stories. We propose a computational pipeline that automatically extracts a story's temporal narrative verb-based event chain for each of its characters as well as character attributes such as gender. We also present a verb-based event annotation scheme that can facilitate bias analysis by including categories such as those that align with traditional stereotypes. Through a case study analyzing gender bias in fairy tales, we demonstrate that our framework can reveal bias in not only the unigram verb-based events in which female and male characters participate but also in the temporal narrative order of such event participation.

%% file: 1_introduction.tex
Social biases and stereotypes are embedded in our culture in part through their presence in our narratives \cite{taylor2003}. %
Despite the focus on documenting and mitigating the social bias that arises from the %
pre-trained embeddings used in 
natural language processing (NLP) \cite{zhao-etal-2018-learning, kurita-etal-2019-measuring, Lu2018GenderBI, sheng-etal-2020-towards}, these methods also lend themselves to analyzing the biases within existing texts \cite{asr2021}. Meanwhile, the humanities and social sciences have a rich history of analyzing social bias in texts such as literary works, news reports, and fairy tales~\cite{garry2017archetypes}. However, these analyses are often conducted manually and at a small scale. Advances in natural language processing now allow for in-depth, large scale analyses of social biases within narrative texts. As storybooks, especially fairy tales, are particularly important to children's mental, emotional, and social development \cite{StorybookDevelopment1990, narahara1998gender} 
, we use fairy tales as our genre of analysis. In this paper, we analyze the gender bias in children's fairy tales by comparing the event chains of female versus male characters. 

 Bias  within the field of NLP can take on many different meanings \cite{blodgett-etal-2020-language}. 
 We adopt \citeauthor{blodgett-etal-2020-language}'s definition of social bias as representational harm through social group stereotypes. These groups can be based on social attributes such as gender, race, economic class, and so on. We focus on gender bias as it is a crucial axis of social bias and has extensive work in the NLP literature, including the comparison of word embedding directions \cite{Bolukbasi2016ManIT} and the analysis of the gender representation in literary characters \cite{premodernEnglish2022}. %
 Few studies have considered gender differences in terms of narrative events such as \newcite{Sun2021MenAE} who demonstrated gender differences in celebrity Wikipedia pages by extracting action event triggers. We build upon this work by considering not just event triggers, but chains of event triggers in temporal order. %

A narrative can be simplified into a sequence of events in which a character participates as an agent (the entity which carries out the event) or as a patient (the entity onto which the event is done) \cite{kroeger2005}. %
By considering the sequence, or chain, of events of characters, we can analyze the story narrative in greater detail. To accomplish this task, we develop a data processing pipeline which automatically extracts the temporal narrative event chains of characters, the characters' gender, and the characters' thematic roles in the event.
We group events into event types to simplify analysis and focus on categories of interest which follow historical gender stereotypes.

In summary, our paper presents three main contributions
:
\begin{itemize}
\item We develop a pipeline\footnote{Our Python library (NECE: Narrative Event Chain Extraction Toolkit) which implements the pipeline is open-source and available for download at \href{https://ibm.biz/fair-fairytales}{https://ibm.biz/fair-fairytales}.} for extracting characters, characters' attributes (such as gender), narrative events chains, and characters' involvement in the events as agents or patients from narrative text.
\item We design an event annotation scheme and dictionary for verb-based events that accounts for limitations in existing verb clustering schemes such as WordNet \cite{wordnet} and VerbNet \cite{verbnet}.
\item We demonstrate the first results, to our knowledge, of temporal event chain differences between female and male characters (as agents and patients) in a narrative text corpus through the case study of fairy tales.

\end{itemize}

%% file: 2_related_work.tex
\subsection{Traditional Approaches to Social Bias in Narrative Text}
Traditionally, the analyses of social stereotypes and bias in narrative have been the realm of the social sciences and humanities including literary studies \cite{LitGen1996}, feminist and gender studies \cite{haase2000feminist}, race and ethnicity studies \cite{RaceSFLeonard2003}, queer studies \cite{greenhill2018sexualities}, pedagogy \cite{TeachingReading2013}, and so on. The examination of gender in literature spans across various genres and formats such as classical Greek literature \cite{Greek1995}, news articles \cite{RacePress1991, cdaPoliticsSriwimon2017}, science-fiction \cite{grasf2015}, and early American literature \cite{RaceAmericanLitSundquist1998}. 

One common method to examining these themes in narrative is content analysis, a systemic technique that identifies and groups units in text into categories based on explicit coding rules \cite{stemler2000}. These units can be as simple as words which are quantitatively measured using word frequencies. The units can be more complex, such as themes, which can cover words, phrases, sentences, or paragraphs within a text. Results can be quantitative or qualitative in nature such as reports of frequencies or discussion of identified patterns. Another common interdisciplinary approach is critical discourse analysis \cite{fairclough2010cda} which aims to explain assumptions about the power relations between social identity through the analysis of linguistic features in text. While such approaches allow for in-depth analyses of the text, they require extensive manual coding in order to extend results beyond a small number of specific works. 

\subsection{Gender Bias in Fairy Tales}
The analysis of gender bias in fairy tales is particularly salient as storybooks are important to the development of children's self image and understanding of the world \cite{narahara1998gender, StorybookDevelopment1990}. This includes fairy tales' power to harm children's self image through the perpetuation of harmful stereotypes \cite{hurley2005seeing, block2022}. While fairy tales were originally meant for adult or general consumption, in modern times they were re-framed as children's stories that institutionalized power relations including gender roles \cite{zipes1994, politics1994} and thus make-up one of the largest and "longest existing genres of children's literature" \cite{hurley2005seeing}. 

The analyses of fairy tales has a rich history in social science literature. Since the 1970's, feminist scholarship has debated the benefit \cite{lurie1970liberation} and harm \cite{lieberman1972someday} of the representation of women in fairy tales, with more recent scholarship acknowledging the complexity of such representations \cite{haase2000feminist}. Critical discourse analysis, as described above, has also been applied to fairy tales to investigate the relationship between the powerful and the powerless \cite{shaheen2019exploring}. Taylor presents a teaching lesson for conducting content analysis of gender stereotypes in children's books \cite{taylor2003}.

\subsection{Natural Language Processing Approaches to Social Bias in Narrative Text}
Much of the existing work in social bias in natural language processing is concerned with detecting and mitigating the bias of language models \cite{zhao-etal-2018-learning, kurita-etal-2019-measuring, Lu2018GenderBI, sheng-etal-2020-towards}. For example, the word embeddings used in many of these models can be shown to be biased towards a particular gender, such as "homemaker" towards "woman" and "programmer" towards "man" \cite{Bolukbasi2016ManIT}. Such analyses are necessary  but limited, especially when trying to capture more nuanced biases in existing narrative texts beyond correlations between words. Traditional social science and humanities approaches are more suited to capturing nuance but have their own drawbacks as discussed above.

To overcome the limits of manual coding, researchers have begun to leverage other NLP methods to analyze bias in narratives at scale. NLP methods lend themselves particularly well to content analysis as they automate the counting of text units such as words, characters, and semantic relations. For literary texts, \newcite{premodernEnglish2022} use a common NLP method (Named Entity Recognition), a sequence comparison library, and a gender detector library to extract characters and their genders with the goal of comparing the number of female and male characters that appear in pre-modern English literature. Their results show that male characters appear far more often than female characters at a rate of 8 to 5 which reflect the results of similar studies using manual coding \cite{mccabe2011childrensbooks}.
Crucially, we follow \newcite{Sun2021MenAE}'s use of odds ratios as our gender bias metric. In analyzing the career and personal sections of celebrities in the Wikipedia corpus, they find that women's marriages were more often linked with their careers while men's marriages were considered part of their personal history instead. %
This paper extends prior research by examining gender bias not only in individual events but also in the sequence of the temporal ordering in which they occur, providing a more comprehensive analysis of the issue.

%% file: 3_data.tex
For our analysis corpus, we used the FairytaleQA dataset~\cite{xu-etal-2022-fantastic}, which contains 278 open-source fairy tales downloaded from Project Guttenburg. This corpus was originally compiled to train question answering models that could be leveraged to help children learn reading comprehension skills~\cite{zhao2022educational,yao2021ai}. The corpus includes many popular fairy tale collections such as the Brothers Grimm, The Green Fairybook, and the collected works of Hans Christian Anderson. The fairy tales come from a variety of cultures including German, Chinese, Native American, and Japanese (Table \ref{tbl:cultures} in Appendix \ref{a:3}). The average length of the stories is 2,533 tokens. The shortest story has 254 tokens and the longest has 8,847 tokens. 

\begin{figure*}[th]
\includegraphics[width=1.0\linewidth]{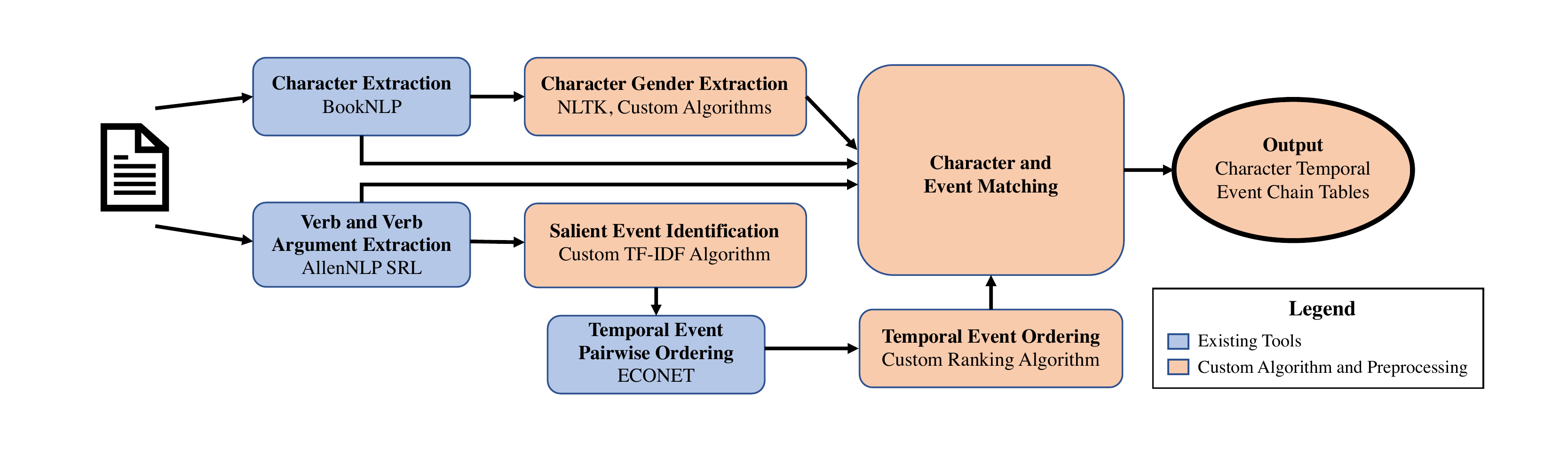}
\caption{Character and Event Extraction Pipeline}
\label{fig:pipeline}
\end{figure*}

\subsection{Character and Event Chain Extraction Pipeline}
In order to analyze the gender bias in narrative event chains of fairy tales, we developed a data processing pipeline (Figure \ref{fig:pipeline}) to extract key narrative features such as main characters, gender attributes, verb events and their temporal order, and salient events of the plot. More specifically, we leverage BookNLP's ``Big'' model \cite{Bamman2014ABM} to extract characters through their character clustering and co-reference resolution algorithms; we improved BookNLP's main character identification algorithm by counting not only direct name mentions of the character, but also pronoun mentions of that character. We defined main characters as those that appeared at least 67\% as often as the character with the most appearances. We developed our character gender prediction models based on pronouns in the co-reference chains as well as gendered words in the character names. Characters whose gender was not specified were classified as ``uknown''. We used AllenNLP Semantic Role Labeling \cite{Gardner2017AllenNLP} to extract verbs along with their subjects and direct objects which served as the triggers for our events. To filter out auxiliary verbs and generic events not important for narrative, we designed a salient events identification model based on the tf-idf algorithm. Lastly, we use ECONET \cite{Han2021ECONETEC} to predict the pairwise temporal relationships between two events. We developed a ranking algorithm to create sequential event chains for all characters based on the pairwise ordering results from ECONET. For more information on these customized algorithms, see Appendix \ref{app:alg}. For all existing models, we ran the models using the default settings and parameters.

\subsection{Extraction Pipeline Validation}

The quality of the event chain from the 
pipeline was assessed by human evaluation of the temporal event ordering and feature extraction components. 

\begin{table}[t]
\centering
\resizebox{0.5\textwidth}{!}{%
\begin{tabular}{llll}
\toprule
Event Chain Detection            & Accuracy   & Macro-F1 & N \\ \midrule
Event Salience          & 0.734     & 0.721        & 188         \\
Character-Event Relationship    & 0.872     & -        & 188         \\
Character Gender        & 0.974     & 0.951       & 188       \\ \bottomrule
\end{tabular}
}
\caption{Evaluation of Pipeline Feature Extraction. Note: Only accuracy is reported for character resolution because number of character classes is not fixed across different stories.}
\label{tab:eval_feature_extraction}
\end{table}

For the temporal ordering evaluation, we asked annotators to rank extracted verb events from a given passage into sequential temporal order. We compared these ranks with  Kendall's $\tau$ coefficient, which measures the similarity of the orderings of the data~\citep{Kumar2010GeneralizedDB}. The result was a Kendall's $\tau$ coefficient of 0.974. The high performance can be explained in part by the high quality temporal model of ECONET and in part by the relative simple narrative structure of fairy tales in which most events follow a sequential order. 
For feature extraction, evaluators annotated 188  sentences from 11 stories across the three dimensions as shown in Table~\ref{tab:eval_feature_extraction}. 

Annotators were asked if the extracted verb event was important to understand the main plot of the story. They were then asked to identify the relationship between an extracted character and the extracted verb event: agent, patient, both agent and patient, or not related at all. Lastly, they were asked to infer the gender of the extracted character. We imagine that the evaluation of the salient event detection scored relatively low (F1 of 0.72) in part because of the high subjectivity of the task especially given insufficient prior examples. However, we do believe there is definite room for improvement of the salient event detection algorithm. Meanwhile, the character-event relationship and character gender extraction algorithms perform very well (F1 of 0.87 and 0.97 respective) because of the high quality of the BookNLP and AllenNLP pipelines. Overall, the robust results from our integrated, developed pipeline lend us confidence in using extracted event chains to perform our bias analysis.

Overall, the robust results from our developed pipeline lend us confidence in using extracted event chains to perform our bias analysis.

%% file: 4_annotations.tex
There has been substantial previous work in annotating and clustering verbs. BookNLP \cite{Bamman2014ABM} clusters event entities into nine supersense categories such as \emph{body}, \emph{communication}, \emph{competition}, \emph{emotion}, and \emph{possession} based on WordNet's lexicographer files \cite{wordnet}. VerbNet \cite{verbnet} clusters events into many of the same categories but includes more fine-grained groups to cover a total of 101 types and 270 classes. However, the categories from these two sources are not immediately useful for our analysis as the categories tend to include both synonyms and antonyms. For example, the event ``harm'' is categorized in the sub-class ``amuse'' in VerbNet along with events such as ``please'', ``comfort'', ``delight'', and ``encourage''. Given the subject of our analysis, there were also some important missing categories related to common male and female stereotypes such as a grouping of domestic tasks or actions common in battle. To address these limitations, we used a mix of automated and manual methods to annotate the event types.

\subsection{Annotation Process}

We first used automated methods as a starting point for our event type annotations. The first step in grouping events was to lemmatize verbs to a single word. For instance, the verbs ``say'', ``says'', ``saying'', and ``said'' are grouped as ``say''. We matched each lemmatized verb to its BookNLP supersense category, VerbNet class, and VerbNet sub-class. 
Then, we manually checked the three categories for each lemmatized verb. Of all the verbs, 21\% were not found in VerbNet and had to be manually matched to a category. We tended to default to the more fine-grained VerbNet classes over the BookNLP supersense categories. Overall, about 30\% of events retained their VerbNet class and sub-class. For verbs that were grouped with their antonyms, we created a new class or sub-class such as the class ``harm''. We also created new classes to capture the common stereotypes such as women being associated with domestic labor (``clean'' and ``cook'') and men being associated with business and achievement. In addition, new sub-classes helped distinguish broad classes; the ``domestic'' class was given sub-classes of ``clean'', ``cook'', ``decorate'', and so on. Around 24\% of verbs were re-categorized into these new classes and sub-classes over those of VerbNet. Meanwhile, 11\% of the verbs were originally grouped into a VerbNet class and/or sub-class that included antonyms and so were also re-categorized. 

One major limitation was that our pipeline does not determine the semantic meaning of the extracted verb. Thus, polysemous verbs could be matched with multiple, often unrelated classes. In cases where we found that the word overwhelming had a single meaning in the fairy tale corpus, we matched it with a single class and sub-class. Otherwise, we did not match the event with any class. Polysemous verbs accounted for 7\% of all verbs. 10\% of the verbs were not matched with any category because the most common meaning could not be established or because the verb did not fit into any of the defined categories. Ultimately, we decided on 97 classes and 172 sub-classes which are listed in detail in  Table \ref{tbl:annotations} in the Appendix.

\subsection{Historically Stereotyped Event Types}
Out of our 97 classes we picked out 16 classes (see Table \ref{tab:select_event_classes}) that aligned with traditional gender stereotypes. Many of these corresponded to the adjectives used by Taylor \citeyearpar{taylor2003} in their male and female coding frames. Feminine descriptions included submissive, unintelligent, emotional, passive, and attractive. Masculine traits included intelligent, rational, strong, brave, ambitious, active, and achievement. We also referenced the Personal Attributes Questionnaire, a 24 item questionnaire that was intended to measure gender identity by linking gender identity to common gender stereotypes such as women to crying, the home (domesticity), and helpfulness and men to aggression, competition, and determination \cite{paq1975}. The newly created classes extending VerbNet are shown in bold in Table \ref{tab:select_event_classes}.

\begin{table}
\begin{center}
\begin{tabular}{ cc } 
 \toprule
 Female & Male \\
 \toprule
\textbf{emotion} & \textbf{knowledge} \\ 
\textbf{passive} & \textbf{active} \\ 
\textbf{submissive} & \textbf{obstinate} \\ 
 helping & \textbf{authority}  \\ 
 \textbf{domestic} & harming \\ 
 \textbf{intimacy} & \textbf{business} \\ 
 crying & \textbf{success/failure} \\
 & \textbf{battle} \\
& killing \\
 \bottomrule
\end{tabular}
\end{center}
\caption{Selected Event Types by Gender Stereotype. Classes that extend VerbNet are shown in bold.}
\label{tab:select_event_classes}
\end{table}

%% file: 5_methods.tex
Our primary numerical measure of bias is the odds ratio as used in \citet{Sun2021MenAE}. While typically used in fields such as medicine, it can be easily adapted and interpreted in the context of narrative bias. For example, in a given story, the occurrence of the event ``kill'' has an odds ratio of four from male to female characters. This means that male characters are four times more likely than female characters to be involved in an event regarding killing. We apply a common correction, Haldane-Anscombe, to account for cases in which one group has no observed counts of the event \cite{Lawson2004OddsRatio}. To estimate the significance of biases' odds ratios, we calculate 95\% confidence intervals using 1,000 bootstrap samples. We randomly sample, with replacement, 1,000 sets of the 278 stories from the FairytaleQA corpus. Odds ratios are calculated for each event type for each bootstrap sample. If the confidence interval of an event type does not contain 1.0, it suggests that the bias towards that particular gender is statistically significant.

We are also interested in whether a character is the agent or patient of an event. A character is considered the agent (the entity doing or instigating the event) %
if the Semantic Role Labeling model identified them as the subject of the verb event. Likewise, a character is considered a patient (the entity onto which the event is done), %
if the Semantic Role Labeling model identified them as a direct object of the verb event. 

Comparing the event chains of characters is non-trivial. A diverse set of verbs can cover the same event or type of event. The FairytaleQA corpus contains 1,431 unique events, many of which only occur a few times. This scarcity is compounded when considering the chains in which an event occurs as well as whether the character was involved as the agent or patient. Additionally, characters have event chains of different lengths which correlate with character importance to the story. The bias towards male characters appearing more often in fairy tales also means that male characters will tend to have longer event chains. To facilitate analysis, event chains were broken down into segments or normalized. %
We always calculate separate odds ratios for events in which characters were agents or patients. In order to ensure a sufficient sample size, we only considered analysis units (unigrams, bigrams, etc.) that occurred at least five times in the corpus. In summary, we perform three types of analysis:

\begin{itemize}
  \item \textbf{Unigram Event Comparisons:} We compare the odds ratios between female and male characters for single events regardless of position in the event chain. 
  \item \textbf{Bigram Event Comparisons:} Bigrams (chains of two events) are extracted from each event chain. For example, a common bigram is (``communication'', ``travel''.) For each event type anchor \textit{a}, we compare the odds ratios between male and female characters for the event type before and after event type \textit{a}. 
  The most common event types were communication, body movements/motion, travel and so most event bigrams had at least one of such types. Because about 80\% of these were minor, non-salient events like ``say'', ``tell'', ``ask'', ``come', ``go', and ``walk' and to focus on the events most salient to the plot, we filtered these event types from the event chains. Thus a chain of (``communication'', ``harm'', ``communication'', ``communication'', ``emotion'') became (``harm'', ``emotion'').
  \item \textbf{Event Chain Section Comparisons:} To account for the variety in event chain lengths, we normalized the temporal order into the beginning, middle, and end of the event chain for each character. Each section represents one third of the chain and can be compared to the sections of other character chains no matter the chain length. Odds ratios between male and female characters were calculated for an event occurring in each temporal section of the chain. %

\end{itemize}

For an illustrative example of how an event chain is broken up into the above analysis units, please see Figure \ref{fig:analysis_unit_visual} in Appendix \ref{a:3}.

%% file: 6_analysis.tex
The FairytaleQA corpus contained 33,577 events involving male and female characters of which 69\% were attributed to male characters and 31\% to female characters. These events were categorized into 172 event types including a type 'other' for events that do not fit in any other class.

We focused on the event types related to common gender stereotypes shown in Table \ref{tbl:sterotype_events_dist_top_verbs} in the Appendix. 

\begin{figure}[h]
\includesvg[width=1.0\linewidth]{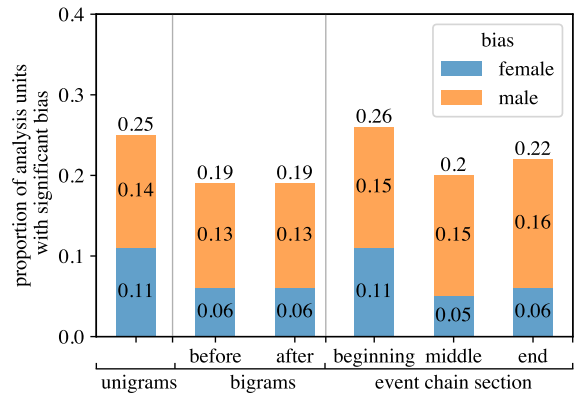}
\caption{Proportion of Significant Gender Bias by Analysis Unit. When not considering temporal order, 1 in 4 event types are gender biased. Temporal differences are represented by the ``bigrams'' and ``event chain section'' groupings. When looking at event bigrams, 1 in 5 show statistically significant bias. When looking at the location of an event within a character's narrative arc, female characters have more biased events in the beginning of their arcs while the bias for male characters is fairly consistent throughout all three sections of their arcs. }
\label{fig:analysis_units}
\end{figure}

\subsection{Event Type Unigrams}

We calculated the odds ratios between female and male characters for the 257 of 293 event sub-class and argument pairs that had at least 5 occurrences in the corpus. Out of these, 14\% of pairs are biased towards male characters and 11\% are biased towards female characters (Figure \ref{fig:analysis_units}).

\begin{figure*}[h]
\includesvg[width=1.0\linewidth]{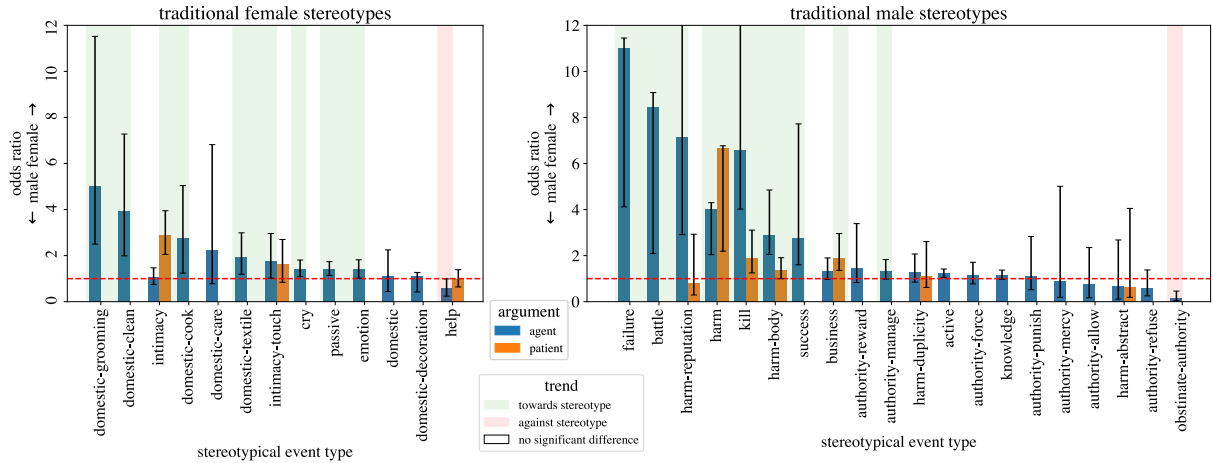}
\caption{Unigram Odds Ratios by Stereotype Event Type. When not considering temporal order, events in fairy-tales show statistically significant differences that typically follow gender stereotypes with a few exceptions.}
\label{fig:stereotypes_uni}
\end{figure*}

When considering the stereotypical events listed in Table  \ref{tbl:sterotype_events_dist_top_verbs} (Appendix), our fairy tale corpus mostly follows these gender stereotypes as seen in Figure \ref{fig:analysis_units}. Many of the top ten events of female (Table \ref{tbl:uni_top_10_f}, Appendix) and male (Table \ref{tbl:uni_top_10_m}, Appendix) characters follow the expected gender stereotypes. The most stereotyped events for female characters were specific domestic tasks (grooming, cleaning, cooking, and textile) while the most stereotyped events for male characters involved events related to failure, success, or aggression. %
We saw smaller, but still significant differences for the passive/active divide. For the emotion/knowledge divide, we only saw small significant differences for female characters for events involving emotions but no significant difference for events involving knowledge. This might be due to our annotation schema being too general in its definition of knowledge events as it includes every instance of ``think''. For some categories, differences depended on the thematic relation of the character. For example, general intimate events like marriage were 2.9 times more likely to have female patients but intimate physical events like hugging and kissing were 1.8 times more likely to have female agents. 

Two event types showed significant results for odds ratios against the expected gender direction. The event type ``help'' (for agents) was biased towards male characters - not female characters as historical stereotypes would lead us to expect \cite{paq1975, taylor2003}. Instead, we find that male characters in fairy-tales are often described as supporting their parents (particularly mothers) or helping someone with a quest. Another event type that went against the historical stereotype was the event of type ``obstinate-authority'' which, instead of being biased towards male characters, was actually 6.8 times more likely for female characters. Indeed, the plots of many fairy-tales that center female characters revolve around the character disobeying her parents or other authority figures; this occurs across cultures such as in the Japanese folktale ``The Bamboo Cutter and Moon Child'' and the Native American folktale ``Leelinau: The Lost Daughter''. This is such a common female plot archetype that the type 'obstinate-authority' has the largest odds ratio for female characters.

\subsection{Event Type Bigrams}
\label{analysis:bigrams}
After removing events of subcategories that were not of analytic interest (``communication'', ``travel'', ``motion'', and ``other'') as well as removing bigrams that occurred less than fives times, we had 327 bigrams of event sub-class and argument pairs such as (harm-body [agent], possession [agent]). When looking at events that happen before a particular anchor event as described in \ref{sec:methods}, 6.4\% show a bias towards female characters and another 13.4\% show a bias towards male characters. When looking at events that happen after particular anchors, 6.4\% show a bias towards female characters and 12.8\% show a bias towards male characters. (See Figure \ref{fig:analysis_units}.) Around one-fifth of all bigrams showed significant gender bias which suggests that gender bias does not only exists for events, but also the order in which the events take place. Many of these bigrams are rather rare even when only considering bigrams that occurred at least five times; 25\% of these occur five times and 75\% occur 11 times or less.

\begin{figure}[ht]
\includesvg[width=1.0\linewidth]{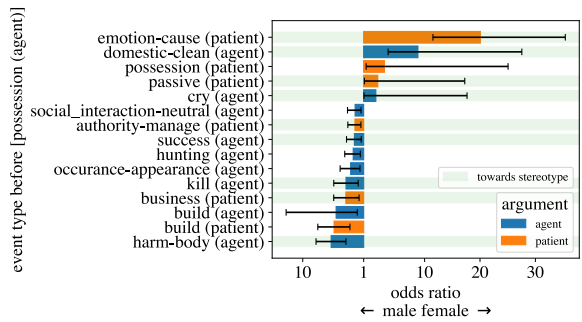}
\caption{Bigram Odds Ratios for Event Types Before Possession (Agent). The difference in previous events suggest that the way in which a character gains or loses possession may be gender biased.}
\label{fig:possession_bigrams}
\end{figure}

\begin{figure*}[h]
\includesvg[width=1.0\linewidth]{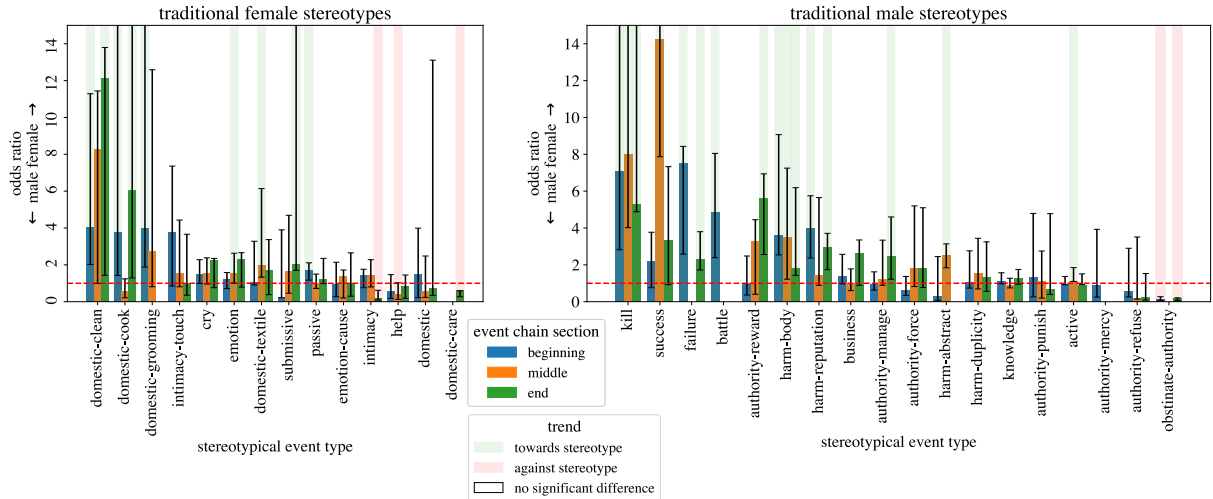}
\caption{Event Chain Section Odds Ratios by Stereotype Event Type. Some event types (such as ``domestic-clean'' and ``kill'') show statistically significant differences towards common gender stereotypes. Other event types (such as ``help'' and ``obstinate-authority'') show statistically significant differences against such stereotypes. This suggests that gender bias exists in \textit{how} a character's narrative arc is structured and not just \textit{what} occurs in such an arc.}
\label{fig:stereotypes_sections}
\end{figure*}

\paragraph{Bigrams with Historically Stereotyped Anchor Event Types.}

Of bigrams occurring at least five times, only fourteen bigrams show significant differences in the event type that happens before a stereotype event. Meanwhile only twenty-one such bigrams show significant differences in the event type that happens after a stereotype event. Nor do the top biased bigrams tend to include as many stereotyped events as the top biased unigrams. (As examples, the top ten biased bigrams for events before the anchor are shown in Appendix Tables \ref{tbl:bi_10_f_before} and \ref{tbl:bi_10_m_before}). This suggests that the greatest gender differences in fairy tale narratives reach beyond our chosen stereotypes. Alternatively, events surrounding stereotype events might be incredibly varied in fairy-tales which makes it hard to access significant differences. We saw evidence for this as many of the bigrams with historically stereotyped anchor event types were too rare to include in our analysis. For example, all bigrams with the event type ``success'' occur less than five times except for the bigram (``success-agent'', ``possession-agent'') which occurs five times.

\paragraph{Non-Biased Event Unigrams with Biased Event Bigrams.}

Some events that were unbiased when considered outside of an event chain showed a gender bias in the events directly surrounding them. For example, the event type ``possession-agent'' showed no significant difference between genders. However, as seen in Figure \ref{fig:possession_bigrams}, many of the events that happen before possession events are gender biased and some of these follow gender stereotypes. (Indeed, many of the events in the top ten most biased bigrams for both female and male characters involved a possession event as shown in Appendix Tables \ref{tbl:bi_10_f_before} and \ref{tbl:bi_10_m_before}.) This difference in previous events suggests that the way in which a character gains or loses possession may be gender biased. This kind of result can encourage researchers to further look into event types or chain combinations that we do not traditionally think of as or expect to be gender biased. 

\subsection{Event Type by Event Chain Section}

When normalizing event chains to beginning, middle, and end character narrative sections, we also find gender differences between female and male characters (as shown in Figure \ref{fig:analysis_units}). The beginning of the event chains appear to have the most female biased events while all sections of the event chain show a similar proportion of male biased events. 

Figure \ref{fig:stereotypes_sections} demonstrates how many of the historically stereotyped event types show strong gender bias in the expected direction across the beginning, middle, and end of a character's event chain. However, the strength of the bias varies by section, and a substantial number of stereotypical event types showed no difference in some of the sections. This suggests that gender bias in events is intrinsically tied to a character's narrative arc structure.

%% file: 7_conclusion.tex
Our character event chain extraction pipeline and odds ratio analysis was able to demonstrate that there are significant differences in not just the events that male and female fairy tale characters participate in, but also gendered differences in the temporal narrative order of such participation. In total, one-fourth of all event types showed significant gender bias no matter the temporal order, one-fifth when considering temporal order of bigram events, and one-fourth when dividing event chains into three equal parts (Figure \ref{fig:analysis_units}). 
This method of analysis offers a more nuanced look at differences in narrative text beyond simply counting the number or appearances of characters by gender or the rate of certain events. The method is supplemented by a more refined event-type annotation schema that separates antonyms and creates new classes that align with traditional gender stereotypes. There is ample room to build upon this analysis with a few distinct possibilities planned for future work. For example, there are numerous alternatives to compare event chains such as expanding the n-gram window or focusing on primary versus secondary characters. The method can be used to compare biases within and across cultural groups and genre. The social biases examined can also be extended by including other social group attributes in the extraction of character attributes such as race and ethnicity, age, and economic class. 
The results of this work further emphasize the urgency that future children-oriented NLP applications such as Storybuddy~\cite{zhang2022storybuddy} should pay extra caution to the potential social biases and stereotypes issues embedded in the data and machine learning models.

%% file: 8_appendix.tex
\subsection{Licensing}

\begin{table*}[h]
\footnotesize
\begin{tabular}{p{2cm} p{1.25cm} p{2.5cm} p{2cm} p{6cm}}
\toprule
Artifact        & Type              & License                                         & Intended Use      & Link  \\ \toprule
FairytaleQA     & Dataset           & Not provided                                    & Not provided      & https://github.com/uci-soe/FairytaleQAData \\
BookNLP         & Software          & MIT (c) 2021 David Bamman                       & Not provided          & https://github.com/booknlp/booknlp  \\
AllenNLP        & Software          & Apache                                          & Not provided       & https://docs.allennlp.org/main/  \\
ECONET          & Software          & Not provided                                    & Not provided      & https://github.com/PlusLabNLP/ECONET \\
VerbNet         & Software, Database & VerbNet 3.2 (c) 2009 by University of Colorado & Not provided & https://verbs.colorado.edu/verbnet/  \\
\midrule

\end{tabular}
\caption{Artifact Licenses and Intended Use}
\label{tbl:licenses}
\end{table*}

\subsection{Customized Algorithms for Extraction Pipeline}
\label{app:alg}
Our extraction pipeline included two customized algorithms for salient event identification and sequential ranking of pairwise temporal event relations.

To filter out AllenNLP extracted auxiliary verbs and generic events not important for narrative, we designed a salient events identification model based on the tf-idf algorithm. The intuition was that events that have unusually high frequency in the target story are often important events for the plot. 

We developed a ranking algorithm to create sequential event chains for all characters based on the pairwise ordering results from ECONET. In circumstances where pair-wise ordering could not disambiguate orders of events, we used the heuristic that events positioned earlier in the passage also happened earlier. We acknowledge that not all events happen in the same temporal dimension and are directly comparable, but we attempted to build a temporal event chain for simplicity of visualizing and interpreting the holistic narrative plot.

\newpage
\subsection{Supplemental Figures \& Tables}
\label{a:3}

\begin{figure*}[th]
\includegraphics[width=1.0\linewidth]{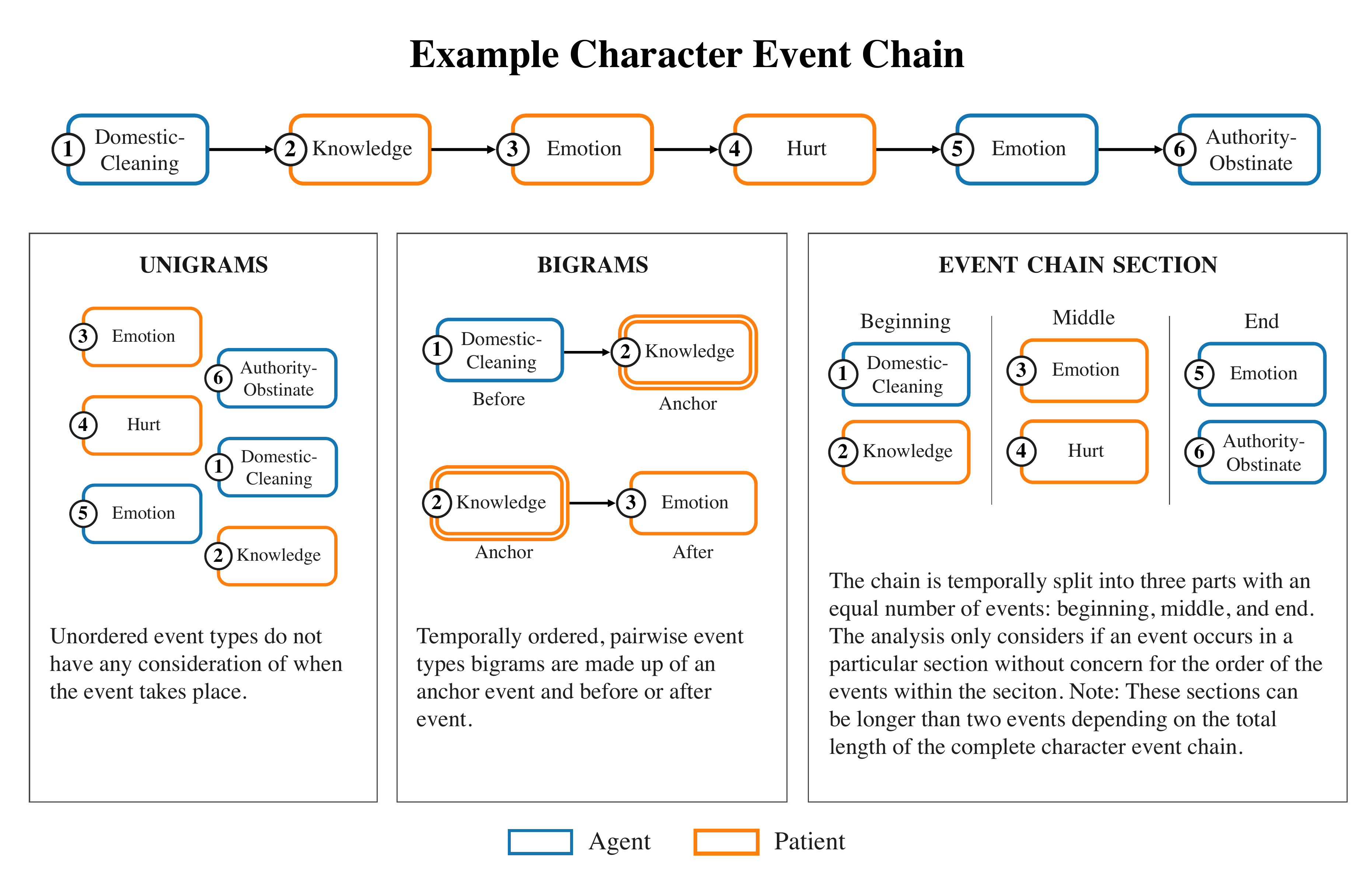}
\caption{Analysis Units Visualization for Example Event Chain}
\label{fig:analysis_unit_visual}
\end{figure*}

\begin{table}[h]
\centering
\begin{tabular}{ll}
\toprule
Culture            & N   \\ \toprule
Scandinavian       & 84  \\
Celtic             & 45  \\
Chinese            & 28  \\
Native-American    & 24  \\
English            & 21  \\
Japanese           & 20  \\
German             & 18  \\
French             & 11  \\
Finnic             &  5  \\
Slavic             &  3  \\
American           &  3  \\
Greek              &  2  \\
Arabic             &  2  \\
Portuguese         &  2  \\
Australian         &  2  \\
West African       &  1  \\
South African      &  1  \\
Romanian           &  1  \\
Spanish            &  1  \\
Indian             &  1  \\  \bottomrule
\end{tabular}
\caption{Distribution of Fairy-Tales in FairytaleQA Dataset by Culture}
\label{tbl:cultures}
\end{table}

\begin{table*}[h]
\centering
\begin{tabular}{llll}
\toprule
Event Type      & Stereotype & N    & Top Verbs                               \\ \toprule
knowledge       & Male       & 1564 & know, think, wonder, understand, learn  \\ 
emotion         & Female     & 358  & like, feel, fear, please, enjoy         \\ \midrule
active          & Male       & 1237 & go, run, walk, rise, hop     \\ 
passive         & Female     & 556  & sit, stand, seat, stray, remain         \\ \midrule
authority       & Male       & 899  & lead, order, declare, allow, refuse     \\
authority, submissive      & Female     & 59   & obey, oblige, comply, behave, abide     \\ 
obstinate, authority       & Male       & 21  & disobey, usurp, resist, rebel, remonstrate     \\ \midrule
harming         & Male       & 695  & shoot, strike, cut, blow, steal         \\ 
helping         & Female     & 224  & help, cure, support, aid, nurse         \\ \midrule
business        & Male       & 403  & bid, pay, buy, sell, owe                \\
domestic        & Female     & 536  & wash, comb, cook, serve, tend           \\ \midrule
success/failure & Male       & 170  & lose, try, seize, win, fail             \\ 
intimacy        & Female     & 468  & marry, love, touch, kiss, hug           \\ 
crying          & Female     & 428  & cry, weep, wail, bewail, bleat           \\ \midrule
battle          & Male       & 14   & subdue, war, vanquish, rout, invade        \\ 
killing         & Male       & 273  & kill, hang, slay, slew, murder          \\  \bottomrule
\end{tabular}
\caption{Stereotypical Event Types Distribution and Top Verbs}
\label{tbl:sterotype_events_dist_top_verbs}
\end{table*}

\newpage
\begin{table*}[h]
\centering
\small
\begin{tabular}{lllll}
\toprule
Event Type              & Thematic Relation & Odds Ratio & 95\% CI       & Top Verbs                                       \\
\toprule
obstinate, authority    & agent             & 6.7        & (2.2, 24.4)   &  resist, disobey, remonstrate \\
harm, scare             & agent             & 5.6        & (2.2, 18.4)   &  frighten, startle \\ 
domestic, grooming      & agent             & 5.0        & (2.5, 11.5)   & comb, brush, clothe, plait, bathe  \\ 
domestic, decoration    & patient           & 3.7        & (1.2, 15.6)   & decorate, adorn, fashion              \\ 
domestic, clean         & subject           & 3.9        & (2.0, 7.3)    & wash, clean, iron, wipe, sweep  \\ 
authority, punish       & patient           & 3.5        & (1.1, 9.5)    & punish, disown, rebuke \\ 
celebrate               & agent             & 3.4        & (2.1, 7.3)    & celebrate   \\ 
dressing                & agent             & 3.00       & (1.1, 11.3)   & wear, dress, don, undress        \\ 
intimacy                & patient           & 3.4        & (2.1, 5.6)    & marry, love                 \\ 
domestic, cook          & agent             & 2.9        & (2.1, 3.9)    & cook, bake, feed, fry \\
\bottomrule
\end{tabular}
\caption{Top 10 Female Unigrams}
\label{tbl:uni_top_10_f}
\end{table*}

\begin{table*}[h]
\centering
\small
\begin{tabular}{lllll}
\toprule
Event Type      & Thematic Relation & Odds Ratio & 95\% CI         & Top Verbs                                    \\ 
\toprule
failure         & agent             & 11.0       & (4.1, 11.5)     & fail, yield             \\ 
bind            & patient           & 10.2       & (2.2, 10.5)     & bind, entrap, mew, wrap                                            \\ 
battle          & agent             & 8.5       & (2.1, 9.1)       & subdue, war, rout, invade, vanquish                                             \\ 
tempt           & patient           & 8.5       & (3.0, 7.9)       & tempt, lure, bait                                             \\ 
engender        & agent             & 7.4       & (2.3, 7.4)       & cause                        \\ 
harm,reputation & agent             & 7.2       & (2.9, 12.2)      & accuse, disgrace, suspect, sue, blame                                         \\ 
harm            & patient           & 6.7       & (2.2, 6.8)       & hurt, harm, maltreat                 \\ 
kill            & agent             & 6.6       & (4.0, 25.0)      & kill, hang, slay, slew                       \\ 
motion, hunting & patient           & 5.8       & (2.1, 6.3)       & ride                          \\ 
motion, forward & agent             & 4.9       & (2.7, 17.3)      & approach, hurry, hasten, advance, chase      \\ 
\bottomrule
\end{tabular}
\caption{Top 10 Male Unigrams }
\label{tbl:uni_top_10_m}
\end{table*}

\begin{table}[t]
\centering
\small
\begin{tabular}{llll}
\toprule
Anchor Event Type                     & Before Event Type                     & Odds Ratio  & 95\% CI \\
\toprule
possession (agent)                    & emotion, cause (patient)              & 34.2        & (13.5, 37.4) \\
social interaction, neutral (agent)   & social interaction, neutral (agent)   & 14.7        & (1.1, 62.9) \\
possession (agent)                    & domestic, clean (agent)               & 12.4        & (5.4, 30.0) \\
intimacy (patient)                    & intimacy (patient)                    & 12.2        & (2.8, 31.5) \\
domestic, clean (agent)               & domestic, clean (agent)               & 7.7         & (2.4, 9.1) \\
passive (agent)                       & build (patient)                       & 7.6         & (2.5, 26.4) \\
want (agent)                          & perception (patient)                  & 7.6         & (2.4, 26.3) \\
emotion (agent)                       & emotion (agent)                       & 6.7         & (3.1, 13.9) \\
send, bring (patient)                 & possession (patient)                  & 6.2         & (1.4, 27.1) \\
social interaction, neutral (patient) & social interaction, neutral (patient) & 6.1         & (1.7, 24.8) \\
\bottomrule
\end{tabular}
\caption{Top 10 (Before, Anchor) Female Bigrams}
\label{tbl:bi_10_f_before}
\end{table}

\newpage
\begin{table}[t]
\centering
\small
\begin{tabular}{llll}
\toprule
Anchor Event Type              & After Event Type       & Odds Ratio  & 95\% CI \\
\toprule
put (agent)                    & possession (agent)     & 13.8        & (4.3, 15.7) \\
possession (agent)             & harm, body (agent)     & 9.5         & (4.2,9.6) \\
possession (agent)             & build (patient)        & 9.1         & (3.5, 9.3) \\
want (agent)                   & find (agent)           & 6.9         & (2.2, 7.8) \\
passive (agent)                & leisure (agent)        & 6.8         & (2.2, 7.9) \\
occurrence, appearance (agent) & possession (agent)     & 6.7         & (2.3, 8.5) \\
domestic, textile (agent)      & perception (patient)   & 6.6         & (2.0, 8.1) \\
possession (agent)             & business (patient)     & 6.3         & (1.8, 6.4) \\
possession (agent)             & kill (agent)           & 6.3         & (2.0, 6.4) \\
passive (agent)                & want (agent)           & 5.7        & (2.2, 6.9) \\
\bottomrule
\end{tabular}
\caption{Top 10 (Anchor, Before) Male Bigrams}
\label{tbl:bi_10_m_before}
\end{table}

\newpage
\subsection{Annotation Scheme}
\label{tbl:annotations}
\bottomcaption{Annotation Scheme}
\begin{supertabular}{p{0.15\linewidth}p{0.15\linewidth}p{0.6\linewidth}}

\toprule
class                   & sub-class       & verbs \\ \midrule
\multicolumn{2}{l}{achievement}           & accomplish, achieve, conquer, defeat, fulfil, fulfill, overcome,   overtake, prevail, relent, succeed, surmount, surpass, surrender, win,   withstand \\
active                  &                 & act, alight, clamb, clamber, climb, crash, crawl, crouch, dandle, dangle,   dart, dash, descend, dismount, drive, fling, gallop, gambol, glide, go, hop, jog, jump, lean, leap, move, plunge, pounce, pursue, race, rise, run,   running, rush, sallied, saunter, skate, skip, slide, soar, speed, splash,   spread, spring, squeeze, step, stick, stray, stride, stroll, swim, swimming, swing, swoop, tramp, tread, trode, trot, vault, venture, wade, walk \\
age                     &                 & age, shrivel, wither \\
\multicolumn{2}{l}{animal sounds}        & bark, buzz, caw, chirp, cluck, crow, growl, howl, quack, roar, snarl, twitter  \\
art                     &                 & draw, paint, perform  \\
art                     & music           & carol, compose, sing, singeth, chant \\
aspectual               & begin           & begin, commence, proceed, start  \\
aspectual               & stop            & cease, desist, end, fade, quit, stop  \\
aspectual               & continue        & continue, repeat, resume  \\
aspectual               & finish          & complete, conclude, finish \\
authority               & manage          & assign, claim, control, decide, declare, destine, direct, dispatch, govern, guide, judge, lead, manage, prescribe, reign, rule, summon, superintend, undertake, usher \\
authority               & punish          & arrest, condemn, confine, disapprove, discharge, dismiss, disown,   persecute, punish, rebuke, suppress, suspend \\
authority               & force           & coax, command, compel, decree, demand, enforce, force, induce, issue,   ordain, order, require, rouse, spur \\
authority               & take            & exact \\
authority               & reward          & anoint, appoint, award, bail, baptize, bless, christen, commemorate,   dedicate, excuse, favor, grant, honor, honour, promote \\
authority               & allow           & allow, permit  \\
authority               & refuse          & decline, deny, forbid, object, refuse, reject  \\
authority               & mercy           & acquit, forgive, pardon, spare, vindicate     \\
battle                  &                 & head, invade, rout, subdue, vanquish, war  \\
bind                    &                 & bind, binding, constrain, entrap, mew, wrap   \\
body                    & touch           & pat, pinch, stroke  \\
body                    & active          & flutter   \\
body                    & putting         & raise    \\
body                    & injury          & bleed    \\
body                    & fear            & flinch, quiver, shake, shiver, shrink, shudder, stiffen, tremble      \\
body                    & sick            & collapse, cough, faint             \\
body                    & submissive      & kneel   \\
body                    &                 & awake, awaken, breathe, curl, knock, pump, roll, shove, slam, spit, stir,   stretch, sweat, wake, waken        \\
break                   &                 & break, destroy, shatter, tear, undo   \\
build                   &                 & assemble, ax, build, carve, construct, dig, erect, fell, fix, forge,   form, frame, hammer, hew, make, making, melt, pave, plaster, repair, saw,   screw, smelt, thatch, weld, wind   \\
business                &                 & afford, apprentice, bargain, barter, bespeak, bid, bribe, buy,   commission, employ, hire, owe, own, pay, profit, purchase, repay, sell, spend    \\
carrying                &                 & carry, drag, haul, heave, hoist, pull, push     \\
celebrate               &                 & celebrate, cheer      \\
change                  & decrease        & crumble, decrease, diminish, dwindle, ebb, lessen, rust, shorten, thin     \\
change                  & stop            & founder, freeze, shut        \\
change                  & positive        & accustom, adapt      \\
change                  & increase        & enlarge, improve, increase, quicken, strengthen, swell   \\
change                  &                 & adjust, affect, alter, balance, become, change, metamorphose, shift,   transform, tweak    \\
choose                  &                 & select  \\
combining               & attach          & attach, band                     \\
\multicolumn{2}{l}{combining}             & bundle, fasten, harness, hitch, join, strap, unite                     \\
communication           & apologize       & apologize, repent    \\
communication           & greet           & greet, hail, wave, welcome  \\
\multicolumn{2}{l}{communication}         & acknowledge, address, admit, advise, agree, allude, announce, answer,   appeal, applaud, appreciate, argue, ascribe, ask, assent, assure, beckon,   begrudge, belabor, bemoan, beware, boast, brag, call, caution, chat, chatter,   communicate, complain, condescend, confess, confirm, congratulate, consent,   consult, contradict, converse, couch, describe, disclose, discourage,   discuss, dissuade, exaggerate, exclaim, explain, express, extol, flatter,   grumble, heed, hint, indicate, inform, insist, introduce, invite, jeer,   mention, mumble, murmur, mutter, name, note, persuade, pledge, praise,   proclaim, profess, promise, pronounce, quote, recite, recommend, recount,   relate, relay, remark, remind, repine, reply, report, reproach, reprove,   retort, said, say, says, scold, scream, screech, shout, shriek, spake, spat,   speak, stammer, state, suggest, swear, talk, talking, tease, tell, thank,   threaten, thunder, utter, whisper, yell \\
communication           & ask             & beg, beseech, enquire, entreat, grovel, implore, inquire, petition,   plead, query, question, request, solicit, urge  \\
consume                 & fast            & fast \\
consume                 &                 & devour, digest, dine, drink, eat, eating, lick, munch, nibble, pour,   quench, sip, suck, sup, swallow, taste   \\
consume                 & dine            & breakfast \\
copy                    &                 & imitate   \\
create                  &                 & conceive, contrive, create, invent, produce, render   \\
cry                     &                 & bawl, bewail, bleat, cry, moan, sob, wail, weep   \\
curse                   &                 & beshrew, curse, haunt   \\
die                     &                 & die, perish  \\
dirty                   &                 & dirty, soil, spoil   \\
domestic                & clean           & burnish, clean, cleanse, dry, dust, iron, polish, purify, scrub, soak,   sponge, sweep, tidy, wash, wax, wipe, wring  \\
domestic                & care            & bandage, calm, care, comfort, console, lull  \\
domestic                & textile         & embroider, felt, knit, lace, sew, shear, spin, stitch, weave     \\
domestic                & cook            & bake, boil, broil, butter, cook, feed, fry, heat, mince, roast, starch, stew    \\
domestic                &                 & attend, entertain, pack, rear, serve, tend, unpack   \\
domestic                & decoration      & adorn, decorate, fancify, fashion, gild, ornament    \\
domestic                & grooming        & bath, bathe, braid, brush, clip, clothe, comb, plait, rinse  \\
dressing                &                 & don, dress, undress, wear  \\
duplicity               &                 & disguise, feign, trespass    \\
eat                     &                 & feast   \\
emission                & sound           & clank, clatter, crackle, jingle, rattle, ring   \\
emission                &                 & emit    \\
emission                & light           & blaze, flash, gleam, glisten, glow, light, shine, sparkle, twinkle   \\
emission                & air             & puff     \\
emotion                 & fear            & dread   \\
emotion                 & cause           & anger, annoy, appease, astonish, bore, delight, disappoint, displease, disturb, excite, fascinate, gratify, heckle, inflame, please, repel, repulse,   satisfy, stun, stupefy, surprise, thrill, torment, transfix, trouble, upset  \\
emotion                 &                 & admire, adore, brighten, cherish, chill, content, despair, despise, disdain, dishearten, dislike, enjoy, fancy, fear, feel, gnash, grieve, hate, hateth, lament, like, louted, mourn, regret, rejoice, relish, resent, sorrow, treasure, whine, worry   \\
engender                &                 & cause    \\
existence               &                 & live   \\
failure                 &                 & fail, mistake, yield  \\
farming                 &                 & cultivate, curdle, distil, herd, milk, mow, pasture, plant, rake, reap,   sow, spade, thresh, unharness, unyoke, water, weed     \\
find                    &                 & discover, examine, find, nose, uncover    \\
forbid                  &                 & bar   \\
forget                  &                 & forget, miscall, mislay    \\
free                    &                 & release   \\
gamble                  &                 & bet, chance, wager   \\
guess                   &                 & assume, guess, presume   \\
harm                    & duplicity       & befool, betray, blindfold, cheat, confound, confuse, deceive, distract,   fool, hoax, lie, outwit, perplex, poison, pretend, rob, snatch, spy, steal, vex   \\
harm                    & scare           & daunt, frighten, startle, terrify  \\
harm                    & abstract        & ail, banish, deprive, detain, harass, imperil, offend, revenge, wrong \\
harm                    & reputation      & accuse, berate, besmear, blacken, blame, disgrace, expose, indict, insult, mock, profane, shame, sue, suspect, upbraid   \\
harm                    & body            & abuse, assail, attack, beat, behead, bite, blow, bruise, burn, butt, choke, claw, cleave, crack, crush, cuff, cut, disfigure, gnaw, gore, hit, inflict, injure, pain, pelt, pierce, prick, punch, scratch, sever, shin,   shoot, slap, sling, smack, smash, smite, spear, squash, stab, sting,   stricken, strike, suffocate, trample, whip, wound, wrestle   \\
harm                    &                 & harm, hurt, maltreat, molest, overpower  \\
harm                    & reptutation     & scorn   \\
help                    &                 & aid, assist, avail, benefit, better, bolster, counsel, cure, heal, help, helping, mend, nurse, revive, support, warn  \\
hold                    &                 & chain, clasp, contain, hold, restrain \\
hunting                 &                 & catch, fish, halloo, hunt, mount, rein  \\
\multicolumn{2}{l}{incompetence}          & droop, flounder  \\
intimacy                & touch           & fondle, kiss, pet, tickle, touch  \\
intimacy                &                 & betroth, caress, embrace, hug, love, marry, nuzzle, wed \\
\multicolumn{2}{l}{investigate}           & investigate, review, test  \\
kill                    &                 & execute, hang, kill, massacre, murder, slaughter, slay, slew  \\
\multicolumn{2}{l}{knowledge}             & ascertain, bethink, concentrate, consider, contemplate, determine, fathom, imagine, inscribe, instruct, interpret, ken, kens, know, larn, learn, lecture, meditate, memorize, muse, plan, ponder, read, realise, realize,   reckon, reflect, study, suppose, teach, think, thinking, understand, wist, wonder, write  \\
leisure                 &                 & amuse, banter, bask, chuckle, dabble, dance, disport, fiddle, frolic, hum, jest, joke, laugh, play, prance, waltz, whistle  \\
lodge                   &                 & quarter, shelter \\
measure                 &                 & enumerate   \\
mistake                 &                 & sin   \\
motion                  & flee            & abandon, avoid, depart, desert, dodge, escape, evade, flee, retreat, shy, slink, withdraw \\
motion                  & hunting         & ride  \\
motion                  & linger          & tarry  \\
motion                  & body            & arch, bow, flap, fly, kick, thrust   \\
motion                  & hide            & conceal, cover, hide  \\
motion                  & forward         & advance, approach, ascend, charge, chase, hasten, hurry, launch, near, outstrip  \\
motion                  & passive         & drift, fall, hover  \\
motion                  & sailing         & moor, row, sail, sink  \\
motion                  & duplicity       & creep    \\
motion                  & putting         & lift, load, lower, shoulder   \\
motion                  & submissive      & follow  \\
motion                  & incompetence    & fumble, hobble, lag, limp, scramble, slip, stagger, stumble, totter, trip, trudge, trundle, tumble  \\
need                    &                 & need   \\
neglect                 &                 & forsake, neglect  \\
\multicolumn{2}{l}{nonverbal\_expression} & blink, blush, flush, gasp, salute, shrug, yawn  \\
nonverbal expression   & negative        & groan, scowl, sigh, sneer, snort     \\
nonverbal expression   & positive        & beam, grin, nod, smile, wink  \\
obstacle                &                 & burden, foil, hinder, interfere, interrupt, prevent, stifle  \\
obstinate               & authority       & depose, disobey, oppose, rebel, remonstrate, resist, usurp    \\
occurrence               & occurrence      & befall     \\
occurrence               &                 & happen, occur    \\
occurrence               & appearance      & appear, arise, burst, emerge, open, reappear    \\
occurrence               & disappearance   & disappear, vanish   \\
participate              &                 & partake, participate   \\
passive                 &                 & betide, deserve, encounter, experience, float, idle, miss, pace, pause, remain, retire, seat, sit, stand, standeth, starve, stay, stood, struggle, suffer    \\
\multicolumn{2}{l}{perception}            & behold, descry, espy, eye, gaze, glance, glimpse, goggle, hear, listen, look, notice, observe, overhear, peep, peer, perceive, recognise, recognize, scent, see, sense, smell, stare, watch, witness   \\
\multicolumn{2}{l}{perseverance}          & bear, endure, persevere, persist, preserve  \\
possession              &                 & accept, acquire, adopt, allot, attain, bequeath, bestow, borrow, capture, choose, choosing, collect, deliver, devote, dispose, distribute, earn, endow,   exchange, fetch, furbish, gain, gather, get, give, givin, grab, hand, have,   inherit, keep, lack, lend, loan, lose, obtain, offer, pocket, possess,   procure, provide, provision, receive, redeem, regain, retain, reward,   sacrifice, secure, seize, seized, seizing, share, supply, take, taketh, taking, waste    \\
practice                &                 & exercise, ply, practice, practise, train  \\
predict                 &                 & foresee, foretell, predict, prophesy   \\
prepare                 &                 & prepare    \\
prosper                 &                 & bloom, flourish, grow, prosper   \\
\multicolumn{2}{l}{protection}            & accompany, defend, escort, free, guard, protect, rescue, safeguard, save, ward   \\
put                     &                 & arrange, bury, cram, dump, fill, heap, install, pile, place, prop, put,   scatter, set, sprinkle, strew  \\
religion                &                 & pray, pray'd, worship  \\
remember                &                 & recollect, remember   \\
remove                  & hunting         & skin    \\
remove                  &                 & clear, empty, omit, remove, rid, wrest     \\
respect                 &                 & esteem, respect, reverence      \\
rest                    &                 & recline, rest, resteth, sleep, snore, sprawl  \\
sailing                 &                 & capsize, maroon  \\
search                  & hunting         & track   \\
search                  &                 & search, seek  \\
send                    &                 & send   \\
send                    & bring           & bring   \\
separate                &                 & disentangle, divide, part, separate, unfasten, untie  \\
show                    &                 & brandish, display, evince, exhibit, show  \\
social interaction     & combative     & avenge, challenge, compete, dispute, fight, quarrel, spar    \\
social interaction     & neutral         & hobnob, meet, mingle, visit   \\
submissive              & authority       & abide, behave, comply, obey, oblige   \\
tempt                   &                 & attract, bait, bewitch, enchant, entice, lure, tempt    \\
throw                   &                 & pitch, punt, throw, toss    \\
tire                    &                 & exhaust, fatigue, pant, tire, weary    \\
travel                  & leave           & betook, decamp, leave, leaving   \\
travel                  &                 & emigrate, encamp, explore, journey, march, roam, sojourn, transport,   travel, wander, wend   \\
travel                  & arrive          & arrive, come, enter, land, reach, return     \\
trust                   & positive        & believe, depend, entrust, trust      \\
trust                   & negative        & disbelieve, doubt, misgive   \\
try                     &                 & attempt, bestir, endeavor, intend, strive, try    \\
use                     &                 & apply, exert, use  \\
value                   &                 & prize, value   \\
wait                    &                 & anticipate, await, bide, wait  \\
want                    &                 & crave, desire, dream, hanker, hope, long, pine, prefer, want, wish   \\
warm                    &                 & befriend, encourage, gentle, inspire, pity, reassure, relieve   \\
work                    &                 & busy, man, toil, work \\
\bottomrule
\end{supertabular}

\twocolumn